\documentclass[lettersize,journal]{IEEEtran}
\usepackage{amsmath,amsfonts}
\usepackage{algorithmic}
\usepackage{algorithm}
\usepackage{array}
\usepackage[caption=false,font=normalsize,labelfont=sf,textfont=sf]{subfig}
\usepackage{textcomp}
\usepackage{stfloats}
\usepackage{hyperref}
\usepackage{verbatim}
\usepackage{graphicx}
\usepackage{cite}
\usepackage[export]{adjustbox}
\usepackage{graphbox}
\usepackage{color}
\graphicspath{{./figs}}
\DeclareGraphicsExtensions{.pdf,.jpg,.png}
\usepackage{tikz}

\begin{document}

\title{Automatic Fingerpad Customization\\for Precise and Stable Grasping of 3D-Print Parts}

\author{Joyce Xin-Yan Lim and Quang-Cuong Pham
\thanks{J.X.Y. Lim and Q.C. Pham are with the HP-NTU Digital Manufacturing Corporate Lab and the School of Mechanical and Aerospace Engineering, Nanyang Technological University, Singapore.}
}



\maketitle

\begin{abstract}

The rise in additive manufacturing comes with unique opportunities and challenges. Massive part customization and rapid design changes are made possible with additive manufacturing, however, manufacturing industries that desire the implementation of robotics automation to improve production efficiency could face challenges in the gripper design and grasp planning due to highly complex geometrical shapes resulting from massive part customization. Yet, current gripper design for such objects are often manual and rely on ad-hoc design intuition. This would be limiting as such grippers would lack the ability to grasp different objects or grasp points, which is important for practical implementations. Hence, we introduce a fast, end-to-end approach to customize rigid gripper fingerpads that could achieve precise and stable grasping for different objects at multiple grasp points. Our approach relies on two key components: (i) a method based on set Boolean operations, e.g. intersections, subtractions, and unions to extract object features and synthesize gripper surfaces that conform to different local shapes to form caging grasps; (ii) a method to evaluate the grasp quality of synthesized grippers. We experimentally demonstrate the validity of our approach by synthesizing fingerpads that, once mounted on a physical robot gripper, are able to grasp different objects at multiple grasp points, all with tightly constrained grasps.


\end{abstract}


\section{Introduction}\label{introduction}
\IEEEPARstart{T}{he} rise in additive manufacturing comes with unique opportunities and challenges. Massive part customization and rapid design changes are made possible with additive manufacturing, to produce parts used for industrial production or research tasks. However, these parts could consist of highly complex geometrical shapes due to massive customization, which results in challenges posed to manufacturing industries that desire the implementation of robotics automation. A key challenge is the design of the gripper manipulator and the planning of grasps for parts produced by additive manufacturing due to massive customization. This is a valid concern as applying robotics in these areas is an increasing trend~\cite{goel2020robotics}. 

\begin{figure}[t!]
    \centering
	\includegraphics[width=0.35\textwidth]{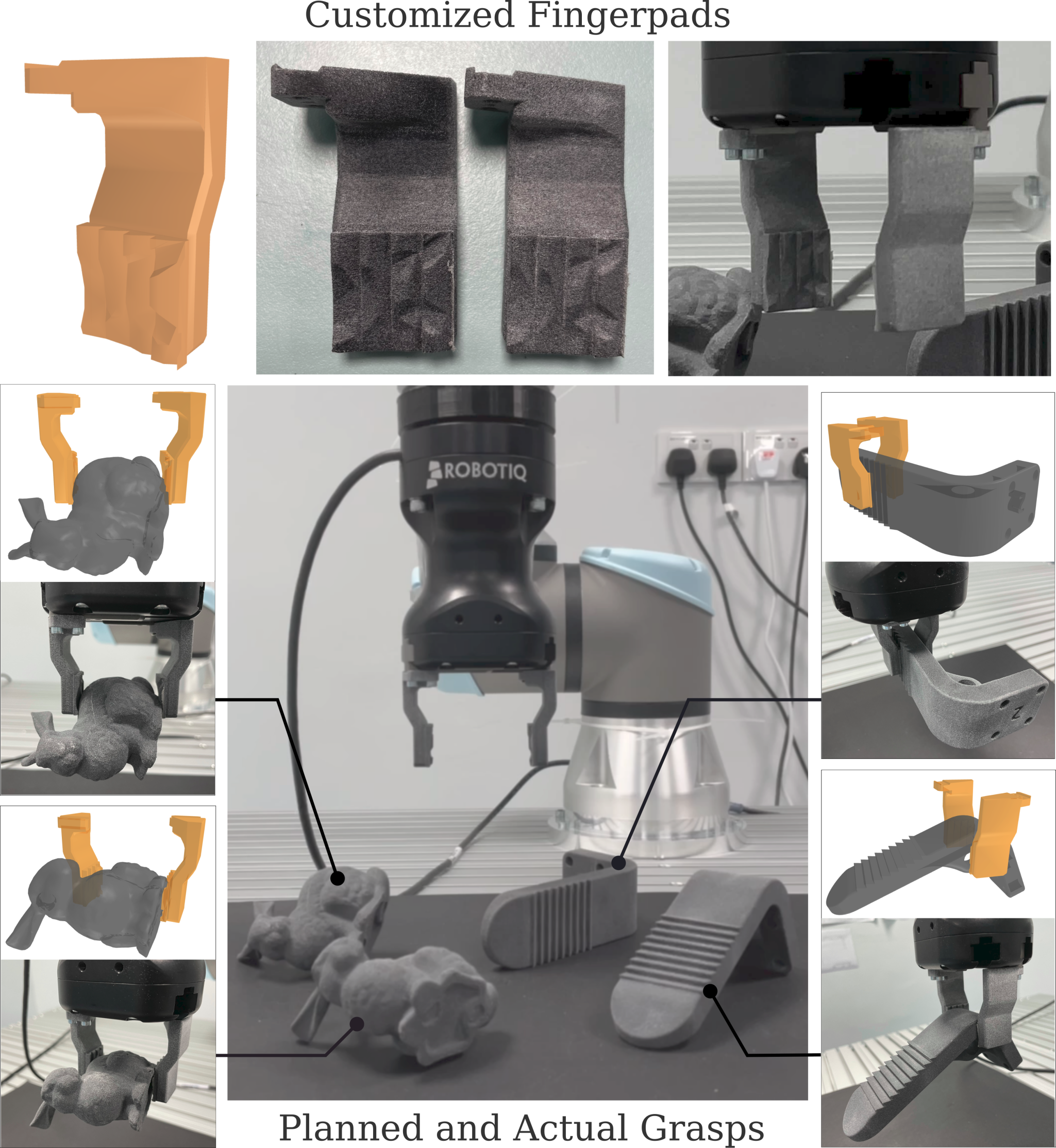}
        \caption{\textbf{Fingerpad Customization with Set Operators
            (FCSO):} A single pair of fingerpads is capable of tightly
          grasping different objects at multiple poses per object. The figure shows a pair of fingerpads that have been designed by FCSO to conform optimally and
          simultaneously to the geometries of four grasped surfaces (2 objects $\times$ 2 poses per object) to form caging grasps. Physical grasping experiments are available at \protect\url{https://youtu.be/M68YagfUF1g}}
        \label{fig:demograsp}
\end{figure}

Current gripper design methods for 3D-Print (3DP) parts are often manual~\cite{fantoni} that rely on ad-hoc design intuition rather than rigorous principles. It would also be difficult for a single manually designed gripper to be able to grasp different complex objects or multiple grasp points of one object. In addition, current automated rigid gripper design processes~\cite{gofd, gafd, phamdt, balan, velasco} tend to produce gripper fingerpads that are based on standard household objects with simple geometries and are also limited to one object at one grasp point. Although these grippers may be able to achieve high-precision grasping due to the close conformation between the fingerpads and the object, they are limiting and not capable of practical implementations due to the lack of versatility~\cite{mahler2018guest}.

An alternative would be to use soft grippers to grasp customized 3DP parts~\cite{9114880, vincent, 6dls, hou2019design, brown2010universal, pacchierotti2017steering}, which are highly versatile but generally lack precision. However, additional techniques would also be required to estimate the pose of the object in grasp. In comparison, customized rigid grippers would generally be tolerant to marginal initial positioning errors~\cite{kaiyu}, because the object in grasp would slide nicely into the geometrical curvature of the gripper that was designed for the particular object pose, which indicates that additional techniques to determine the in-hand pose of the object might not be required.

Thus, due to the challenge posed by massive customization in additive manufacturing, there is a need for a robust, principled method that can automatically design grippers for grasping and manipulating 3DP customized and complex objects, so that automated tasks could be executed. This is also made possible due to the opportunity presented by additive manufacturing that could include the production of these customized grippers for 3DP parts. Hence, we introduce a fast, end-to-end automated approach (Fig.~\ref{fig:demograsp}) to customize grippers for precise and stable grasping of 3DP parts: Fingerpad Customization with Set Operators (FCSO). Our approach relies on two key components:
\begin{itemize}
  \item A method based on set operators (Boolean intersection, 
    union, subtraction), to extract object features and synthesize gripper surfaces that optimally
    conforms to different local shapes: either at different
    grasp points on the same object, or on different objects to form grasping by geometric constraints;
  \item A grasp quality evaluation method for synthesized gripper surfaces to select the optimal gripper surface. This could be an extension of existing grasp indicators, such as force closure~\cite{roboticshandbook, closure1995, gqmeasures}, to emphasize the geometric quality of the grasp for grasps resulting from geometric constraints.
\end{itemize}

\begin{figure*}[htp] 
    \centering
    \includegraphics[width=0.78\linewidth]{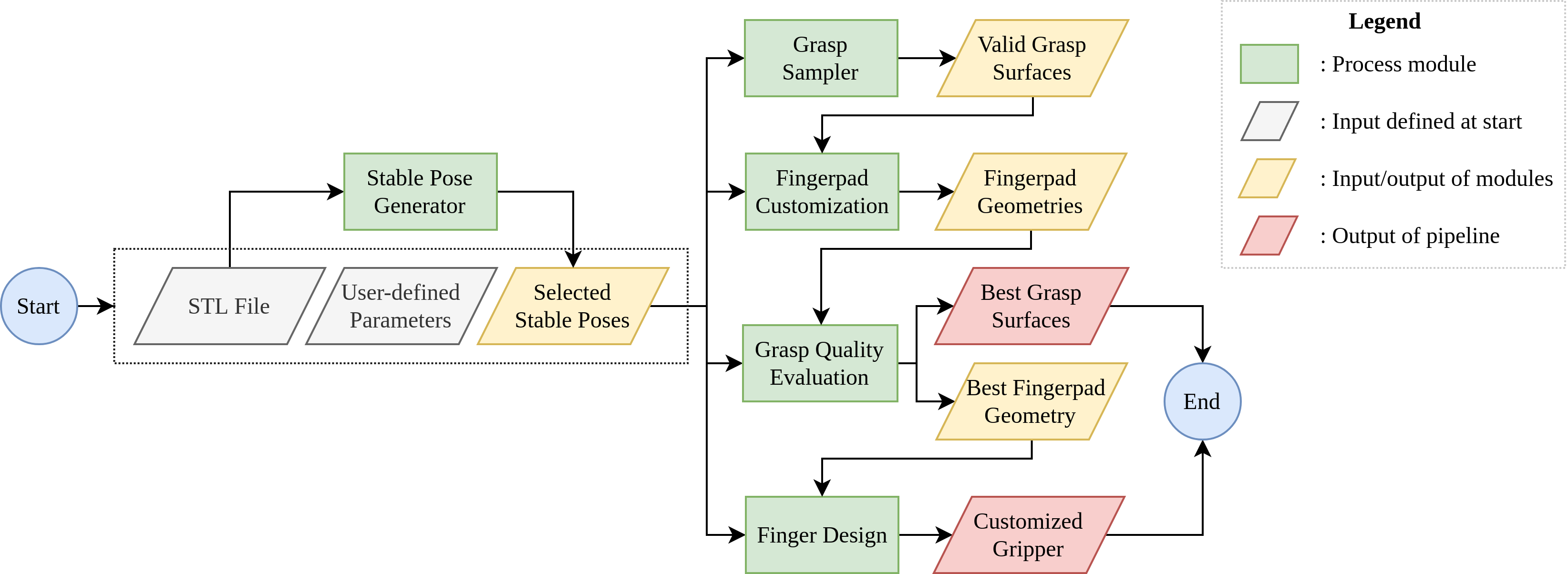} 
    \caption{Proposed pipeline for FCSO. It accepts the STL files of the objects, user-defined parameters from a configuration file, and the flat finger model of a gripper, to automatically return the best grasp surfaces and the best gripper design.} 
  \label{fig:pipeline}
\end{figure*}

We introduce FCSO in Fig.~\ref{fig:pipeline} which consists of five modules: stable pose generator, grasp sampler, fingerpad customization, grasp quality evaluation, and finger design. The stable pose generator accepts the CAD models of objects and user-defined parameters, e.g. gripper specifications, number of stable poses to plan grasps, and size of fingerpad. A set of stable poses, that rest the objects on a planar surface, is returned and stable poses are automatically selected by random. Alternatively, manual selection could be done if specific poses are desired. At each selected pose, grasps are sampled to obtain valid grasp surfaces and locations. Sampled grasps are used in fingerpad customization to extract object features by Set Boolean operators to get fingerpad geometries at each grasp location. The fingerpad geometries at each grasp location are evaluated on their grasp quality to return the best fingerpad geometry and grasp location. The best fingerpad geometry is then fused onto a flat finger to obtain the final customized finger to be mounted on a gripper base.

We aim to leverage the opportunity posed by additive manufacturing to tackle the challenge of gripper design and grasp planning for robotic automation in manufacturing industries that utilize additive manufacturing. The idea is to automatically obtain the design of the gripper fingerpads for a set of objects in one print job, and this is possible as our pipeline only requires the STL files and some user-defined parameters. After printing the fingerpads, which could be printed together with the objects, the fingerpads would be mounted on the gripper and be used in the automation line to grasp and manipulate these objects. The same process can be repeated for a different set of objects in a print job. In the event that the print job is consistent, i.e. for parts that are mass-produced, the same gripper fingerpad can be re-used, and only be printed again after certain wear and tear, which is a generally cheap procedure due to the negligible volume of these fingerpads compared to the actual parts printed for production. As such, the customized fingerpads do not have to be very versatile like soft grippers but require sufficient versatility such as the capability to pick different objects at multiple grasp points so that automated tasks can be practically executed, which is not present in the current state of automated rigid gripper design.

The rest of the paper is as follows: Section~\ref{sec:relatedwork} reviews related literature, Section~\ref{sec:pipeline} introduces the pipeline of our algorithm, Section~\ref{sec:fingerpad_customization} details the method for customized fingerpads. Section~\ref{sec:qualityanalysis} explains the concept for our geometric grasp quality measure and Section~\ref{sec:experiments} evaluates our method from two perspectives through experiments.

\section{Related Work} \label{sec:relatedwork}

\subsection{Rigid gripper customization} \label{ssec:fingercustomization}

Previous studies introduced methods to synthesize gripper
finger designs. In~\cite{phamdt}, appropriate pairs of finger designs are selected from a pre-configured database consisting of simplified geometries. A reconfigurable finger design was proposed in~\cite{balan} that automatically reconfigures cylindrical fingers to obtain a three-point grasp on mainly polyhedral objects. In~\cite{velasco}, a method was proposed to extract the geometry of objects such that fingers generated will enclose the object's surface, forming a grasp by geometric constraint. This method was applied in~\cite{gofd, gafd} where an end-to-end pipeline to obtain customized grippers is limited to individual objects at only one grasp location. 

Recent works for shape optimization include using 3D generative framework~\cite{ha2021fit2form} and graph neural network~\cite{alet2020robotic}. Another study introduced an optimization procedure to cluster geometries and produce a set of robust finger designs that were used to plan grasps of several objects by maximizing contact area~\cite{kaiyu}. However, the size of objects is limited as the center of gravity must reside within the gripper opening. Furthermore,  unfeasible grasps may occur when an object lies exactly in the position of the planned grasp after finger design.

Most studies design grippers for household objects with generally simple geometries such as in the YCB dataset~\cite{ycb}, that may not be applicable to customized objects with complex geometries which are produced by additive manufacturing. These objects could be used in research or industrial production. Our method is robust enough to design grippers for complex geometries with fingerpads that provide geometrical constraints thus achieving secure grasps.

\subsection{Non-rigid grippers}
Soft fingers are versatile as they deform to the local geometry of the object and can better resist external disturbances~\cite{vincent, 6dls}. Studies on hybrid grippers combined soft and rigid structures~\cite{8516330, ma2017yale, patel2021robot} to improve fingertip force, actuation speed, friction, or adaptability. Other soft fingers include conforming pin pads for adaptive grasping~\cite{flintoff2018single} and jamming pads~\cite{hou2019design, brown2010universal} or variable stiffness~\cite{8015159,xu2015design}. However, soft fingers generally lack precision. 

\subsection{Grasp quality measures} \label{ssec:gqmeasures}
Classical point-contact quality measures were discussed in~\cite{gqmeasures} for force closure grasps, including analytical methods using grasp wrench space~\cite{ferrari} that simplifies force closure grasp analysis but cannot take into account the curvature of object’s surface~\cite{caging1}. Recent works on surface-contact quality measures include a surface-contact model that parameterizes the contact area~\cite{s2s}, computation of contact profile using solid geometry intersection and barycentric integration~\cite{reach}, or calculation of contact profile using 6D friction wrench or friction cone~\cite{6dls, 6dfc}. However, simple geometries were used in these works which may be hard to extend to complex contact profiles in customized fingers.

Caging grasps and immobilization are essentially performed based on geometrical constraints~\cite{caging1,caging2}, which make grasps insensitive to friction changes~\cite{kaiyu}. A key idea on two-finger caging grasps is to capture the concavity of the object~\cite{rimon1,rimon2,caging2F} and perform squeezing trajectories~\cite{4209093}, stretching operations~\cite{caging2F} or dispersion control~\cite{4543364} to immobilize objects. A survey~\cite{caging1} discussed types of caging grasps including caging by environment, 2D, 2.5D, 3D caging, and partial caging, however, the authors noted that most practical caging grasps are 2D or 2.5D caging algorithms with parallel jaw grippers. As our gripper customization aims to form caging grasps, we focus on the geometrical constraints and omit friction analysis. We also propose a geometric quality measure to evaluate grasps with only the contact surfaces.


\section{FCSO Pipeline} \label{sec:pipeline}

This section discusses the specifics of FCSO. It accepts the CAD models of objects and user-defined parameters to return a single finger to be mounted on a linear-closing, parallel gripper base. Note that the fingers on the parallel gripper are symmetrical and identical, to illustrate that the pipeline could accept more object poses and geometries.

Detailed explanation of each module is as follows: stable pose generator in Section~\ref{ssec:stablepose}, grasp sampler in Section~\ref{ssec:graspsampler}, fingerpad customization in Section~\ref{sec:fingerpad_customization}, grasp quality evaluation in Section~\ref{sec:qualityanalysis} and finger design in Section~\ref{ssec:fingerdesign}.

\subsection{Specifications} \label{ssec:specs}
We list some specifications used. Software libraries include trimesh~\cite{trimesh} and Blender~\cite{blender}. Specifications of the workstation used are Intel Core i7-6700HQ CPU at 2.60GHz $\times$ 8 with a NVIDIA Quadro M1000M graphics card.

\subsection{Stable pose generator} \label{ssec:stablepose}

The stable pose generator aims to provide several poses that naturally rest the objects on a planar surface, prior to the grasp approach. The stable orientations are estimated with a quasistatic model~\cite{trimesh, goldberg1999part}. The selection of stable poses is random and automatic, and the number of selected stable poses, $N_p$, is pre-defined by the user. If there are specific requirements or prior knowledge on the poses of an object, e.g. for assembly tasks, optional manual input or selection can be conducted. The grasp approach direction is defined in the axis of the world where the gripper approaches the object. A top-to-bottom grasp approach (Z-axis) is chosen by default as side approaches are usually difficult for small objects due to collisions of the gripper base with the table.

\subsection{Grasp sampler} \label{ssec:graspsampler}

\begin{figure}
\centering
\includegraphics[width=.72\linewidth]{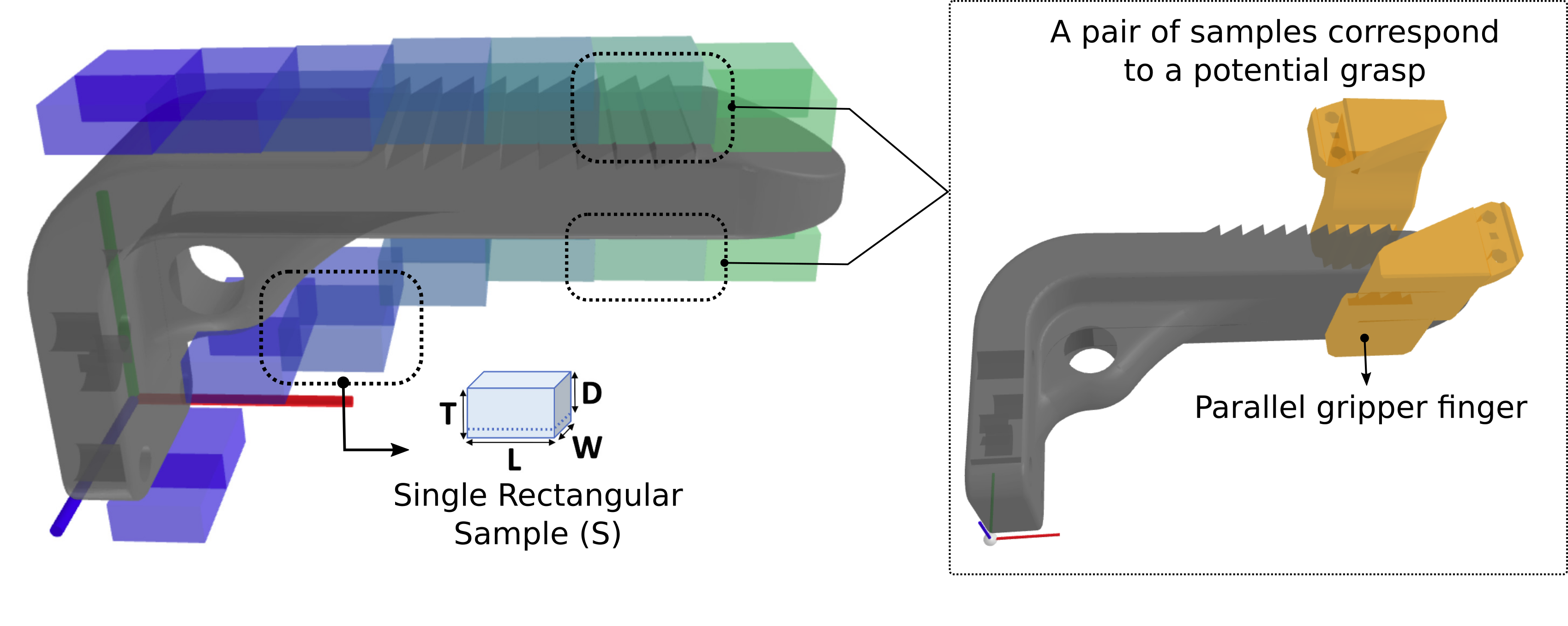}
\caption{\textbf{Grasp sampling} by a sliding pair of rectangular samples $(S)$ along the lateral axis of an object, with a stride equivalent to $L$. Each sample pair has the same color code.}
\label{fig:graspsampler}
\end{figure}

Many tools can plan initial contact locations for basic grippers, such as Graspit~\cite{miller2004graspit} or SynGrasp~\cite{7243306}, or learning-based methods for ambidextrous grasping~\cite{mahler2019learning} and multi-affordance grasping~\cite{zeng2022robotic}. For customized grippers, the local contours are key to forming caging grasps, such as in~\cite{kaiyu, gofd}. Thus, the grasp sampler is required to fully sample the object geometry. It generates candidate grasps by sliding a pair of rectangular samples $(S)$ along the axes of objects, with a sampling step defined as stride~(Fig.~\ref{fig:graspsampler}). This was motivated by the sliding window in neural networks where receptiveness is improved by adjusting the stride~\cite{aloysius2017review}. Similarly, the stride could be applied in grasp sampling to produce more candidates. A smaller stride, or smaller sampling step, returns more grasp candidates. The length $(L)$, width $(W)$ and thickness $(T)$ of $S$ is user-defined. The penetration depth $(D)$ is the amount of penetration of $S$ into the object mesh, and $0<D<T$.

\begin{figure*}[t]
    \centering
    \begin{tabular}{cc}
    \adjustbox{valign=b}{\subfloat[]{\includegraphics[width=.47\textwidth]{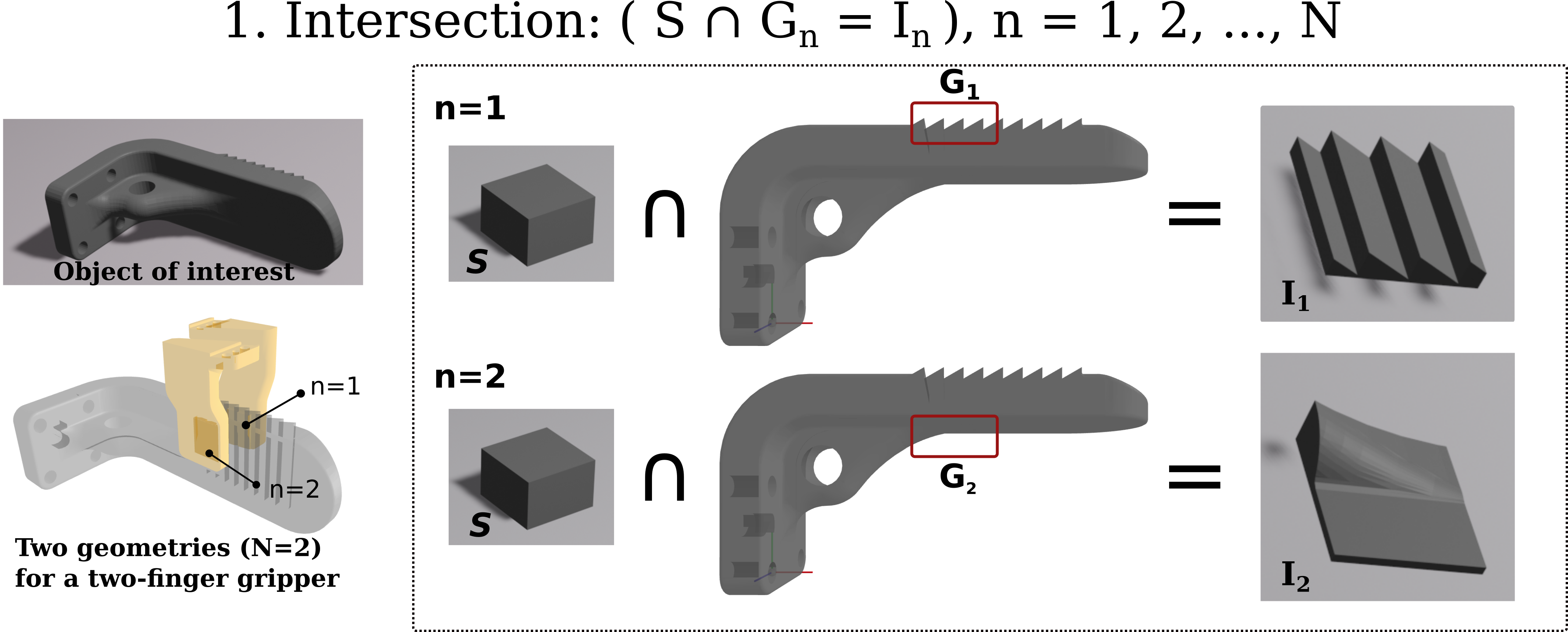}}}
    &      
    \adjustbox{valign=b}{\begin{tabular}{@{}c@{}}
    \subfloat[]{\includegraphics[width=.27\textwidth]{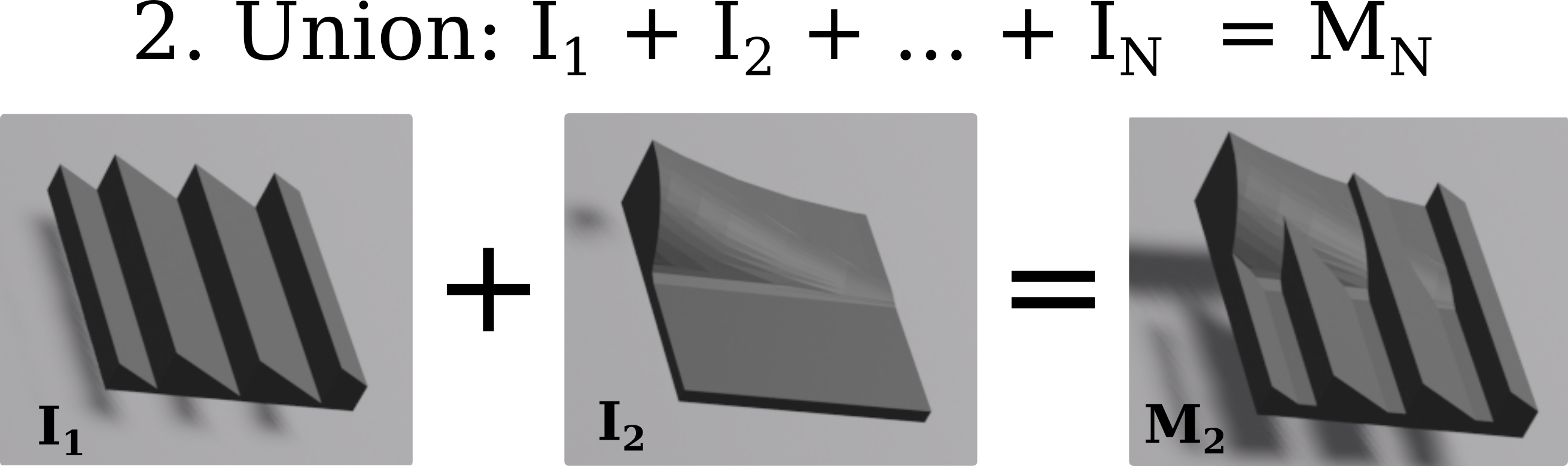}} \\
    \subfloat[]{\includegraphics[width=.27\textwidth]{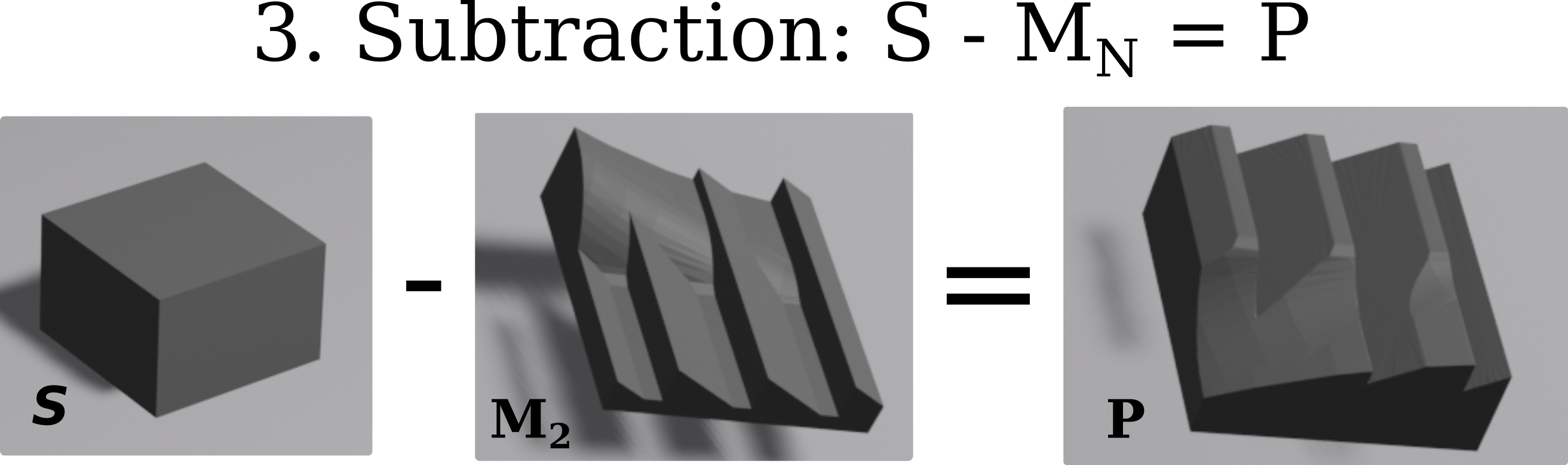}}
    \end{tabular}}
    \end{tabular}
    \caption{\textbf{Fingerpad customization (without filter)} based on the number of geometries $(N)$, while illustrating a three-step procedure on a pair of fingerpads. (a)~Independent Boolean intersections $(I_n)$ resulting from the intersection of every valid rectangular sample $(S)$ and $G_n$, which is the $n^{th}$ geometry of the mesh bounded by the $S$. The samples are obtained from the grasp sampler (Section~\ref{ssec:graspsampler}); (b)~Boolean union of $N$ intersections $(M_N)$; (c)~Boolean subtraction of $S$ and $M_N$ to obtain fingerpad $(P)$ that has a shape which conforms to the mesh at all $G_n$.}
    \label{fig:geometryanalysis}%
\end{figure*}

Feasibility checks are performed on every sampled pair. A sample pair is valid if a sufficiently large contact area can be established during grasping, the grasp is collision-free, and the object can fit into the gripper opening. The number of valid sample pairs for the $m^{th}$ pose is $N_{s,m}$, where $m=1,2,...,N_p$.

\subsection{Finger design} \label{ssec:fingerdesign}

Commercial grippers are often parallel flat finger grippers with basic flat fingerpads,
the CAD model of these basic fingerpads can be retrieved. The optimal
gripper geometry obtained in Section \ref{sec:qualityanalysis} is fused onto the flat finger to obtain the print-ready CAD model of the customized gripper.

\section{Fingerpad Customization with Set Operators} \label{sec:fingerpad_customization}

Caging grasps and immobilization are essentially performed based on geometrical constraints~\cite{caging1,caging2}. Perturbations would not affect the pose of a caged object, thus the pose could be precisely determined with prior information on the gripper. Velasco~\cite{velasco} proposed using Boolean intersections to extract simple, local geometries of objects so that grippers that conform to object shapes can be achieved, but manual grouping is required before subtraction. We extended this concept in our method by using a combination of set Boolean operators with a filter, which allows our method to be sufficiently robust to different object geometries thus achieving an automated design process. Set Boolean operators such as intersections, unions, and subtractions are commutative operations that allow easy addition of new objects or poses.

\subsection{Fingerpad customization without filter} \label{ssec:nofilter}

We define the number of geometries to be extracted as $N$ and rectangular
fingerpad sample, $S$. The $n^{th}$ geometry bounded by $S$ and the mesh is $G_n$, where $n = 1, 2, ..., N$. $I_n$ is the intersection of $S$ with $G_n$ and the union of $N$ intersections is $M_N$. The customized fingerpad is defined as $P$. The method to create $P$ without the automatic filter is shown in Fig.~\ref{fig:geometryanalysis}. This method would generally work well if the sampled geometries are good. An explanation of good geometries is in Section~\ref{ssec:withfilter}.

\begin{figure*}%
\centering\begin{tabular}{c}

\subfloat[]{\includegraphics[width=.38\textwidth,valign=t]{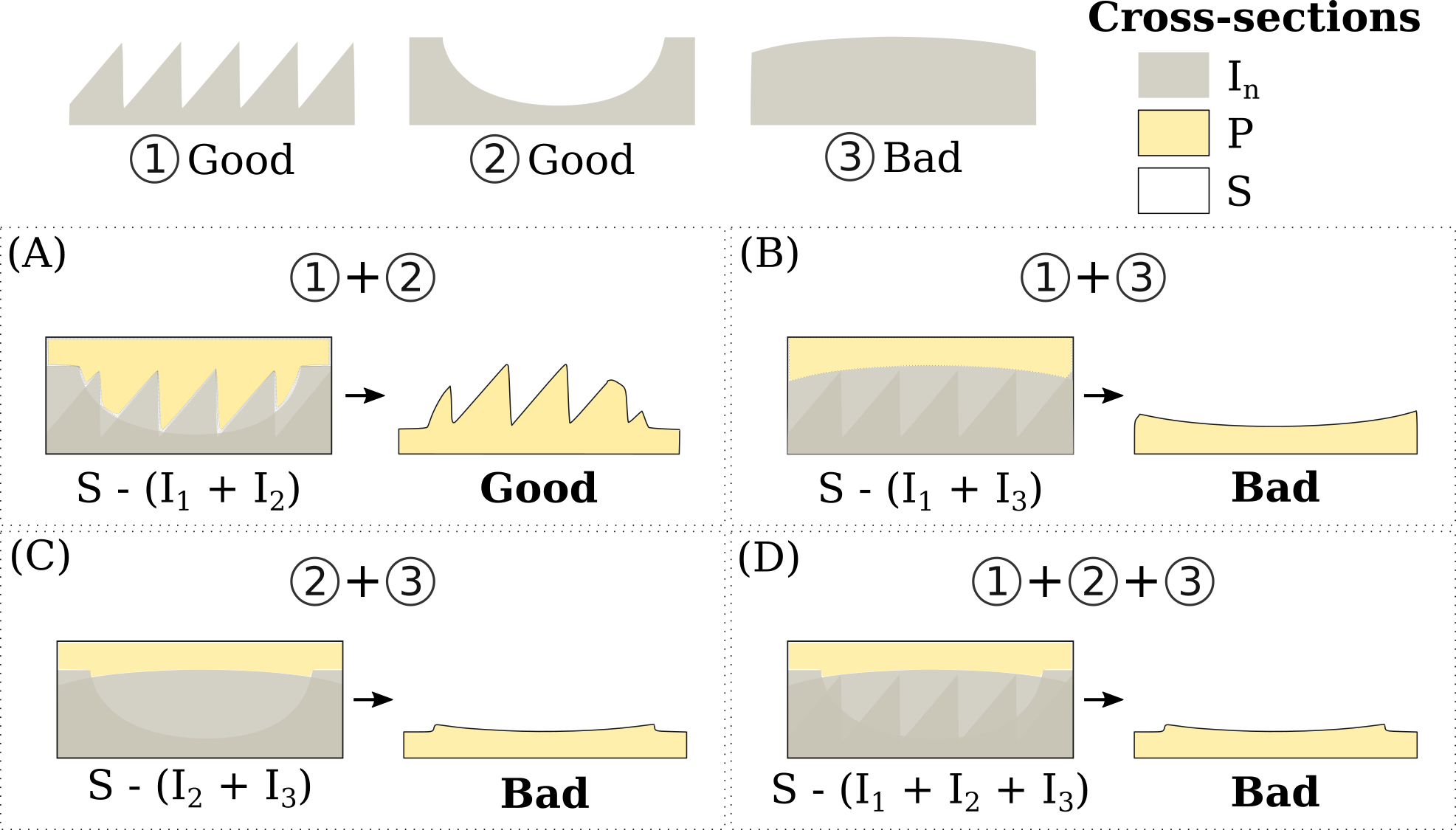}}\hspace{.2cm} 
\subfloat[]{\includegraphics[width=.53\textwidth,valign=t]{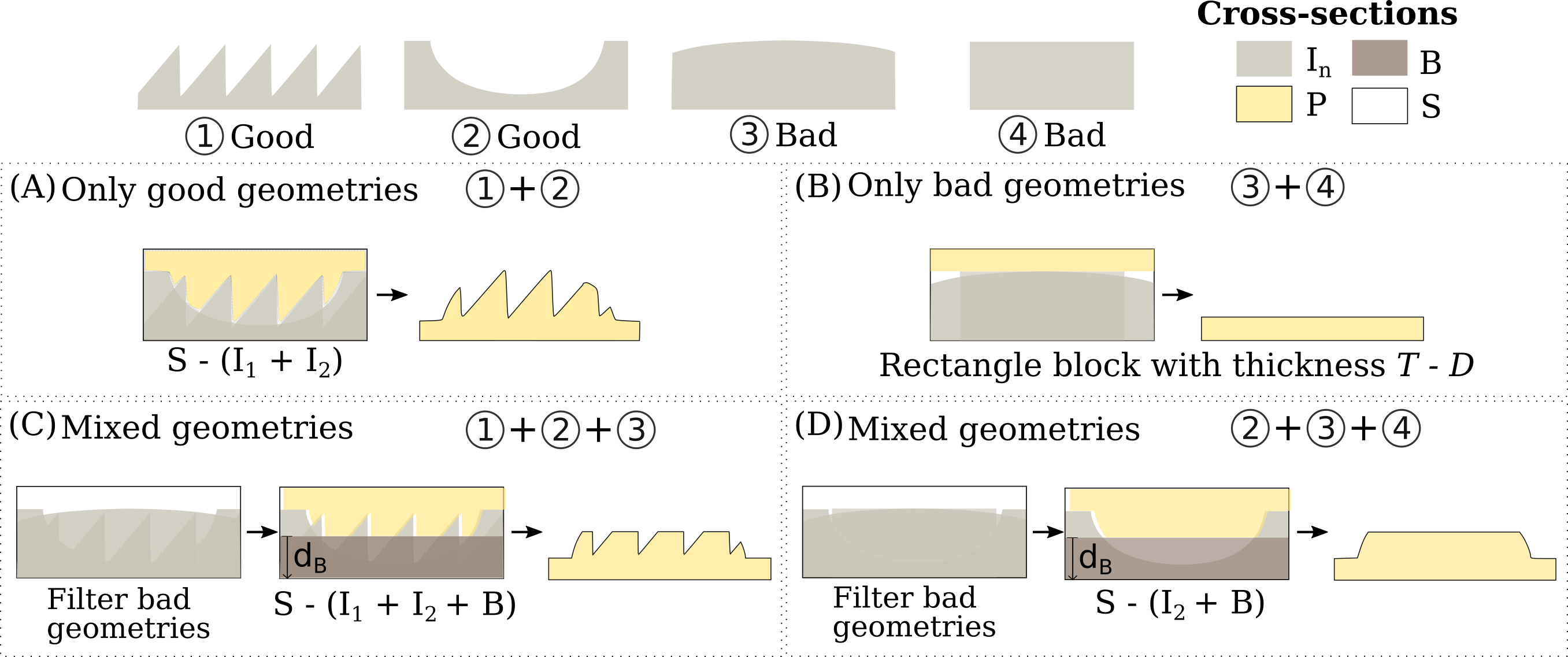}}%

\end{tabular}
\caption{Comparing effects of the filter with good and bad geometries. (a)~\textbf{Without filter}: Undesirable $P$, in yellow, obtained in the presence of a single bad geometry. This shows the need of a filter to differentiate between geometries; (b)~\textbf{With filter}: Visible improved performance. Illustrating three possible cases discussed in Section~\ref{ssec:withfilter}, with $d_1>0, d_2>0, d_3=d_4=0$. In Example C, $d_B=min(d_1,d_2)*K$, whereas in Example D, $d_B=d_2*K$.}%
\label{fig:fcso_filter}%
\end{figure*}

\subsection{Fingerpad customization with filter} \label{ssec:withfilter}
We introduce a volume threshold filter to provide feedback across local geometries. It automatically differentiates `good' and `bad' geometries obtained from set intersections, thus improving the robustness of the geometry extraction to achieve caging grasps. Good geometries are defined as shapes that would create fingerpads that can achieve caging grasps while bad geometries would not achieve such restrictions. The differentiation is crucial as bad geometries such as flat surfaces, are supersets of all geometries, i.e. any geometry $G_n$ can be subtracted from a flat rectangular pad. This also means that any intricate geometries are absorbed by a flat rectangular pad. Thus, if any $I_n$ is flat, $M_N$ would also be flat which results in an undesirable flat fingerpad, $P$. Fig.~\ref{fig:fcso_filter}a shows the absorption of the good geometries in the presence of a single bad geometry. This was avoided with the filter in Fig.~\ref{fig:fcso_filter}b.

\begin{figure}[t] 
\centering\begin{tabular}{c}

\subfloat[]{\includegraphics[width=.65\linewidth,valign=b]{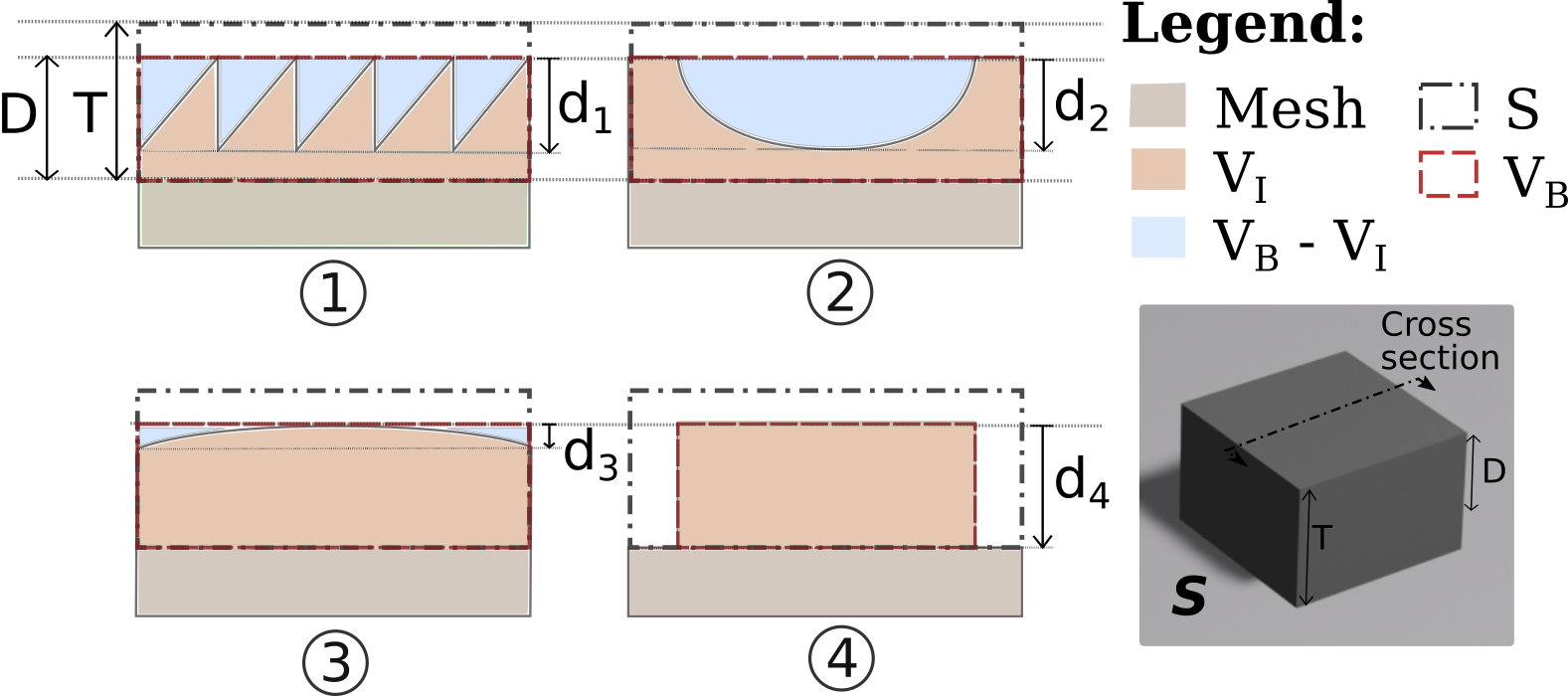}}
\subfloat[]{\includegraphics[width=.34\linewidth,valign=b]{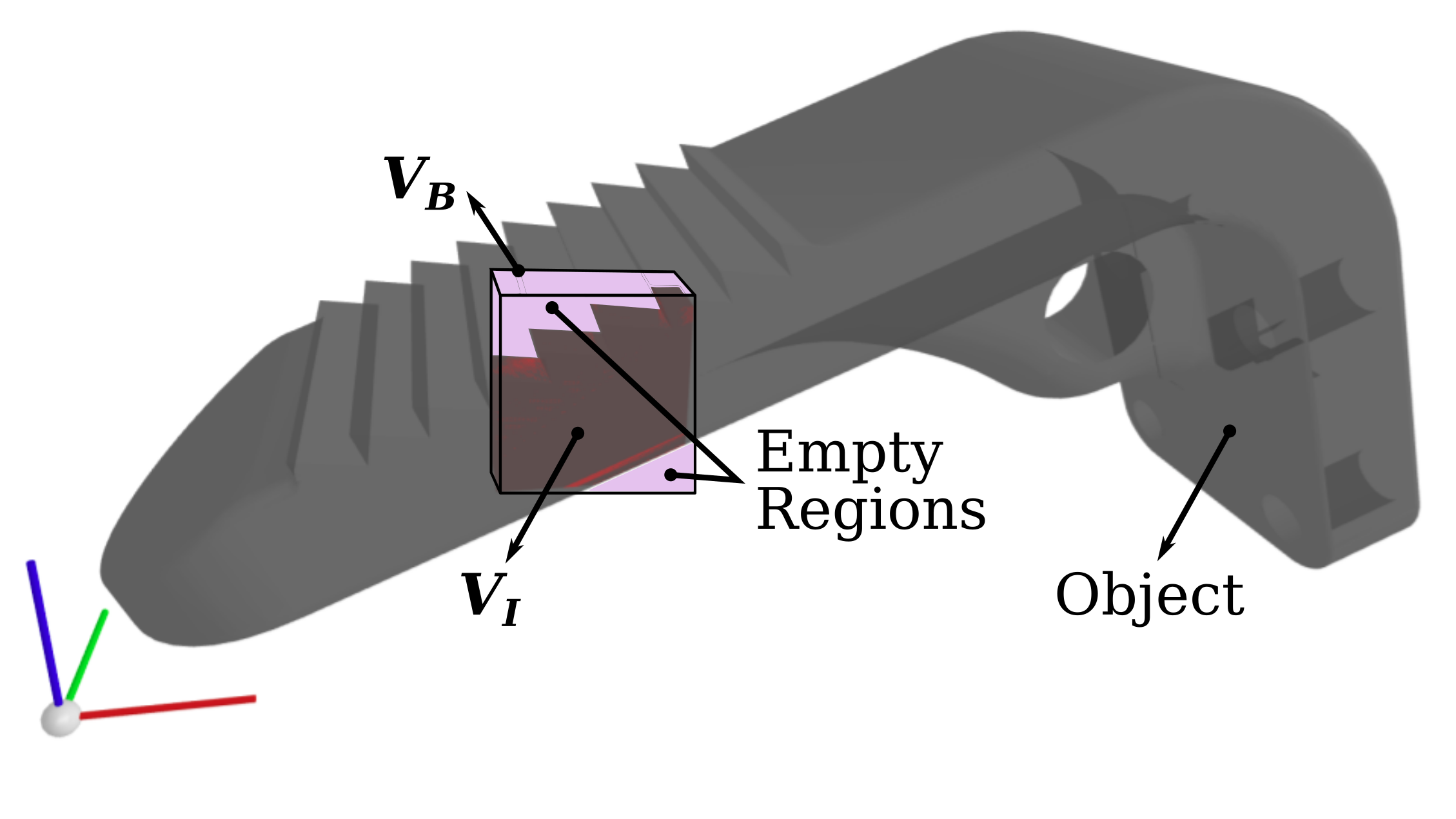}}%

\end{tabular}%
\caption{Volume ratio: (a)Volumes in $R$ and the extracted depth of the geometry of interest $(d)$ in four examples. Note that $V_i$ is a subset of the mesh. Examples 1 and 2 return a large $R$ (good geometries) while Examples 3 and 4 return $R \approx 0$ and $R=0$ respectively (bad geometries); (b) Limitation of volume filter due to empty regions.} 
\label{fig:ratio}
\end{figure}

The differentiation of geometries uses a volume ratio $(R)$ with a constant threshold $(th)$. The volume ratio, $R = (V_B - V_I)/V_B$, where $V_B$ is the bounding box volume of $I_n$, and $V_I$ is the volume of mesh $I_n$. If $R \geq
th$, it indicates that the geometry is good, and if $R < th$, it means that the geometry is bad. This simple yet effective method also filters geometries that are relatively flat, such as edges with fillets as $(V_B - V_I) \approx 0$ which results in smaller $R$s. We suggest using $th=0.1$, which was constant in all experiments of this letter.

A limitation of the volume filter (Fig.~\ref{fig:ratio}) occurs when the mesh edges are at an angle which result in invalid values of $R$. This is due to excess volume in those empty regions of the bounding box, which increases $R$. We require the depth of geometry of interest, $(d_n)$, which is the depth from the object surface to the point where the bounding box of $I_n$ fully encloses the object, to check the validity of $R$. For each $G_n$, if $d_n=D$, any $R\neq0$ is invalid (Example 4 of Fig.~\ref{fig:ratio}). For geometries that lead to invalid $R$, we cluster the surface normals of $I_n$ with similar vector angles. Bad geometries would have the largest cluster perpendicular to the surface of $S$, while good geometries would not. The filtering is complete as every $G_n$ is either labeled as `good' or `bad'.

With the addition of the filter, the creation of $P$ has three possible cases depending on the labels of every $G_n$:

\begin{enumerate}
    \item \textit{Only good geometries:} The three-step procedure in Fig.~\ref{fig:geometryanalysis} executed, resulting in Example A (Fig.~\ref{fig:fcso_filter}b).
    \item \textit{Only bad geometries:} A flat fingerpad
      with a thickness of $(T-D)$ is obtained in Example B (Fig.~\ref{fig:fcso_filter}b). 
    \item \textit{Mixture of good and bad geometries:} For $P$ to achieve good geometric constraints, the first two steps in Fig.~\ref{fig:geometryanalysis} are amended. Intersections are only applied for good geometries and a flat rectangle block $B$ is included during the union to cater for the bad geometries (Examples C and D Fig.~\ref{fig:fcso_filter}b).
\end{enumerate}
The depth of the flat rectangular block $(d_B)$ depends on $d_n$, and $d_n\neq0$ if and only if the geometries are good. As such, $d_B=min(d_1, d_2, ..., d_n) * K$, where $K$ is a constant that affects the degree of `flatness' of $P$. The minimum is considered rather than the maximum so that shallow complex geometries will not be absorbed away by $B$. We suggest using $K=1.5$ which was constant in all experiments of this letter. 

The number of possible fingerpad combinations $(C)$ depends on the number of valid sample pairs and the number of stable placements for planning $(N_P)$. If $N_P=2$ and one pose has three valid pairs of grasp surfaces $(N_{s,1}=3)$ while other pose has four valid pairs of grasp surfaces $(N_{s,2}=3)$, $C=N_{s,1}*N_{s,2}=3*4=12$, meaning that there are 12 possible grippers.

\section{Geometric Grasp Quality Measure} \label{sec:qualityanalysis}

A quantitative measure is needed to evaluate the grasp quality of synthesized gripper surfaces as the caging grasps and immobilization are performed based on a geometrical constraint~\cite{caging1,caging2}, which makes grasps insensitive to friction changes~\cite{kaiyu}. Thus, we propose a heuristic grasp quality measure that emphasizes on the geometric grasp quality.

\subsection{Variation of contact normals} \label{ssec:rles}

In two-finger caging grasps, the concavity of the object is captured to create geometric constraints that immobilize the object~\cite{rimon1,rimon2,caging2F}, which may indicate that the contact surface between the gripper and object, e.g. concave surfaces, has sufficient varying contours that resist perturbations. Thus, a logical heuristic to define geometric grasp quality would be the representation of the variation of contact surface normals, where larger variations of surface normals indicate better grasp.

\begin{figure}[t] 
    \centering
    \includegraphics[width=0.68\linewidth]{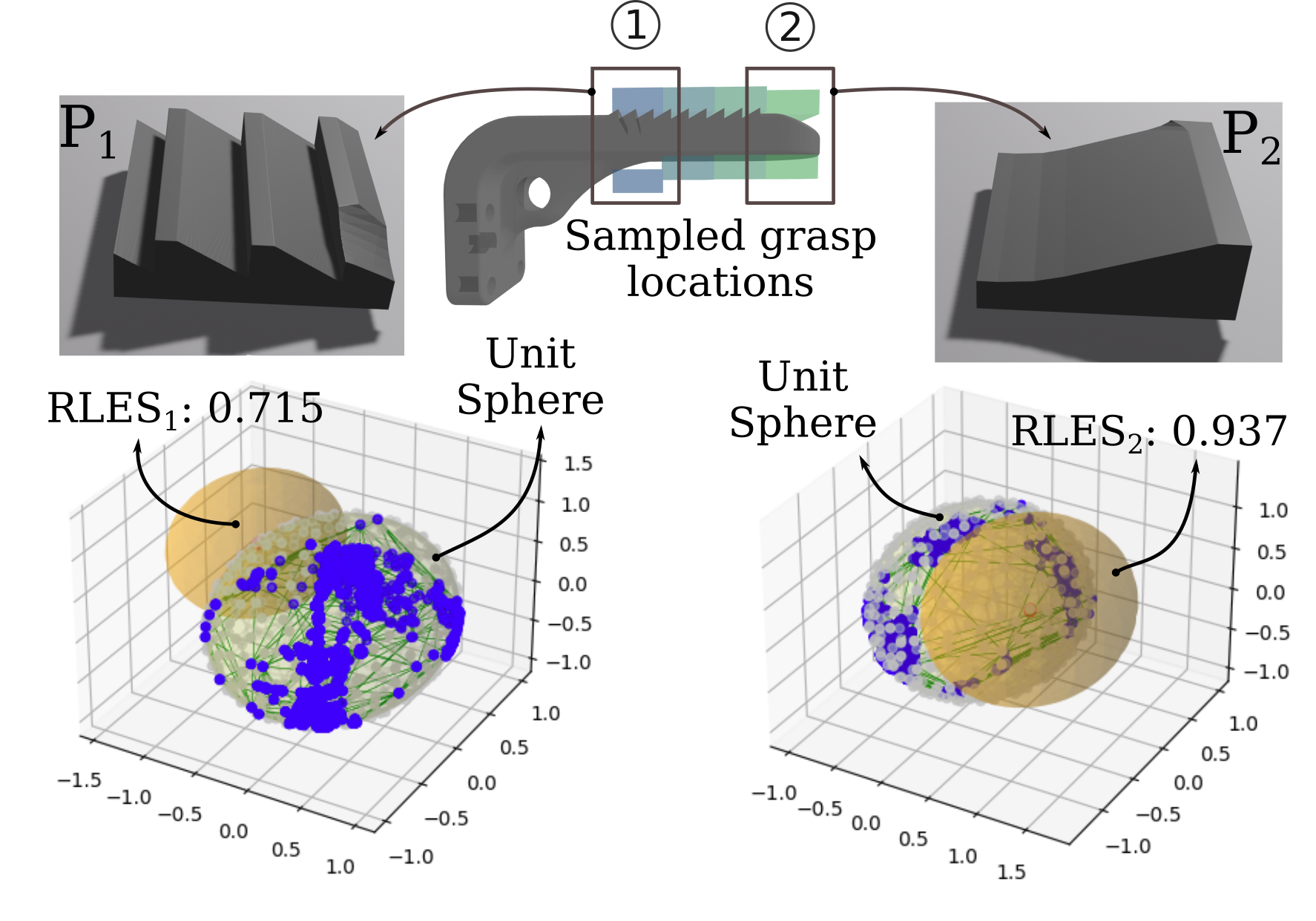} 
    \caption{Quantifying the variation of contact surface normals of fingerpads produced at sampled grasp locations with RLES. Every surface contact normal is mapped to a dot (blue) on a unit sphere. A better grasp would have a larger variation, leading to more dots and a smaller RLES.}
  \label{fig:rlesex}
\end{figure}

The variation is quantified by mapping every surface contact normal of the contact surface between a pair of fingers and a grasped object to a point on a unit sphere (Fig.~\ref{fig:rlesex}), defined as the Radius of the Largest Empty Sphere (RLES). A larger variation of normals would result in a better grasp and denser sphere, which leads to a smaller RLES. Thus, a smaller RLES would indicate a better grasp. It is computed using a combination of 3D Voronoi vertices and Delaunay triangulation. Caroli \textit{et al}
\cite{caroli2010robust} showed that the convex hull of the input
points is equivalent to their Delaunay triangulation on the surface of
the sphere. Megan \cite{schuster2008largest} proposed a solution for
the largest empty circle in 2D by using Voronoi vertices, as the edges
of the Voronoi regions are defined as the circumcenters of the
triangles generated by Delaunay. Hence, the spherical Voronoi vertices
are possible centers of an empty sphere that intersects any Delaunay
triangle at its three ends. A search using KD-trees~\cite{kdtree} is
conducted to compute the RLES.

\subsection{Total surface contact area} \label{ssec:contactarea}
Although the variation of the contact normals may seem sufficient as a grasp quality measure, the total surface area in contact with the object during grasp $(A)$ should also be considered to achieve full geometric constraint, as small grasping areas may cause unstable grasping even with large variations of surface normals. $A$ is the sum of the areas of the finger pair in contact with the object, which is related to the surface normal variation to a certain extent. A larger contact surface would have larger variations if the object is not flat.

\subsection{Quantifying geometric quality of grasps} \label{ssec:graspquality}
Both the variation of contact normals and total contact area are deemed to be equally important. Thus, the effective area $(E)$, is the geometric quality of the $i_{th}$ customized fingerpad at the $m^{th}$ stable pose, by multiplying the inverse of RLES with the total surface contact area at $m$: $E_{i,m}=(1/RLES)*A_m$, where $i=1,2,...,C$ and $m=1,2,...,Np$. A larger $E$ depicts a better quality as it indicates a larger $A$ and better contact normal variation, i.e. smaller RLES.

Each pair of fingerpads is required to grasp object(s) at different pose(s), leading to varying qualities across grasps, i.e. a better grasp may be observed between objects and poses for the same fingerpad pair. Thus, the min-max concept is used, where the quality of the $i^{th}$ fingerpad geometry is the worst possible grasp (minimum $E$) at the $m^{th}$ stable pose: $Q_i=min(E_{i,1},E_{i,2},...,E_{i,m})$. The geometric quality of the best (maximum $Q$) fingerpad geometry is then defined as the $Q_{max}=max(Q_1,Q_2,...,Q_i)$. In simple terms, the grasp quality of each gripper is its worst possible grasp and the best gripper has the highest $Q$ value at its worst grasp across all grippers.

\section{Experiments} \label{sec:experiments}

We evaluate our proposed pipeline from two perspectives:
(1)~Quantitative evaluation of geometric grasp quality measure
(Section~\ref{ssec:gqe}); (2)~Qualitative evaluation of generated
customized fingers for a set of objects and tests of actual
pick-and-place experiments on objects at multiple poses
(Section~\ref{ssec:fingerevaluation}). Note that most objects used were real samples from HP Labs printed for certain industrial tasks. All objects and fingers are printed by the HP MJF5200 using PA11/PA12.

\subsection{Evaluation of geometric grasp quality measure} \label{ssec:gqe}

\begin{figure*}%
\centering\begin{tabular}{c}
\subfloat[]{\includegraphics[height=2.2cm]{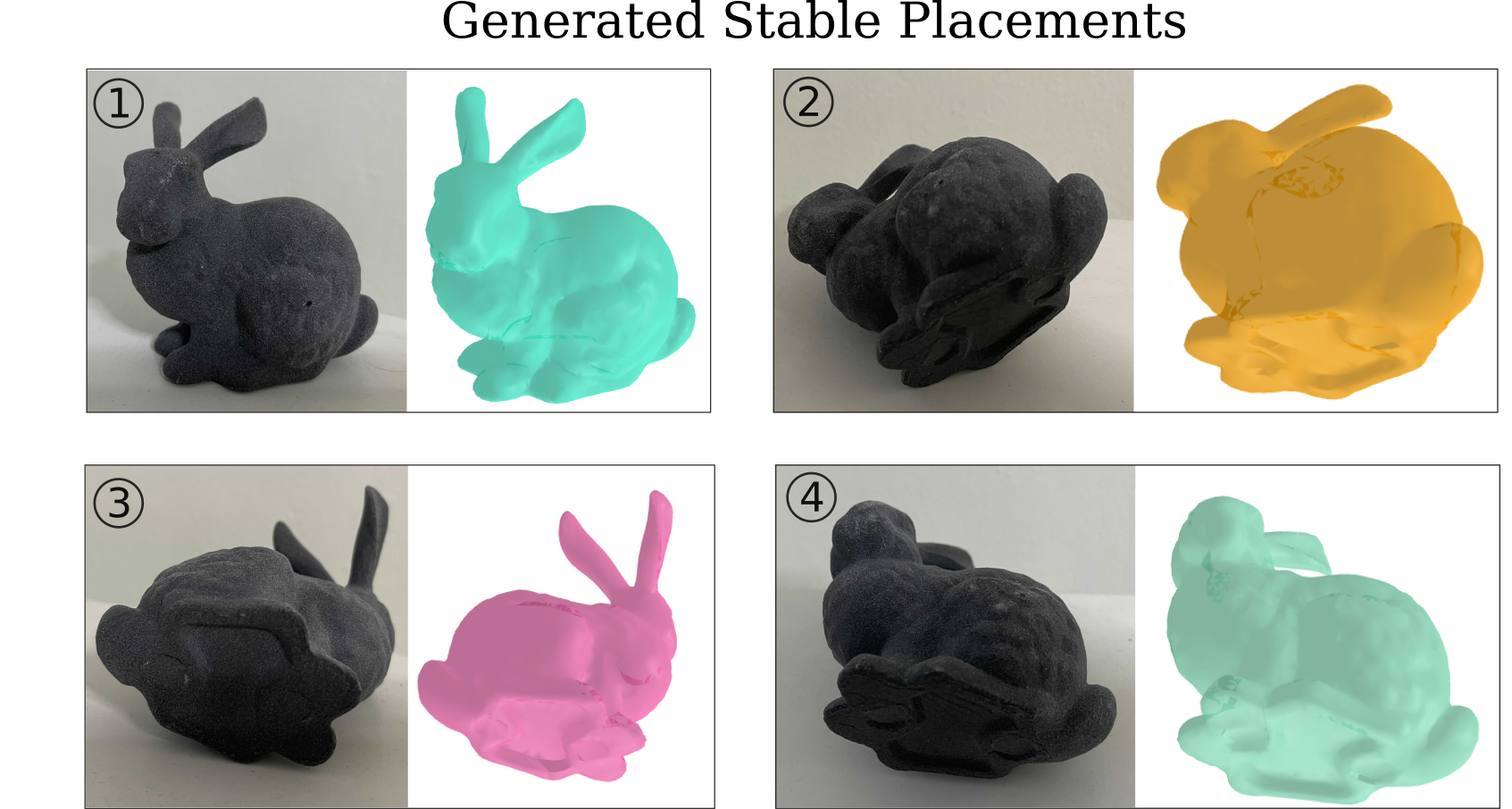}}\hspace{1cm} \subfloat[]{\includegraphics[height=2.2cm]{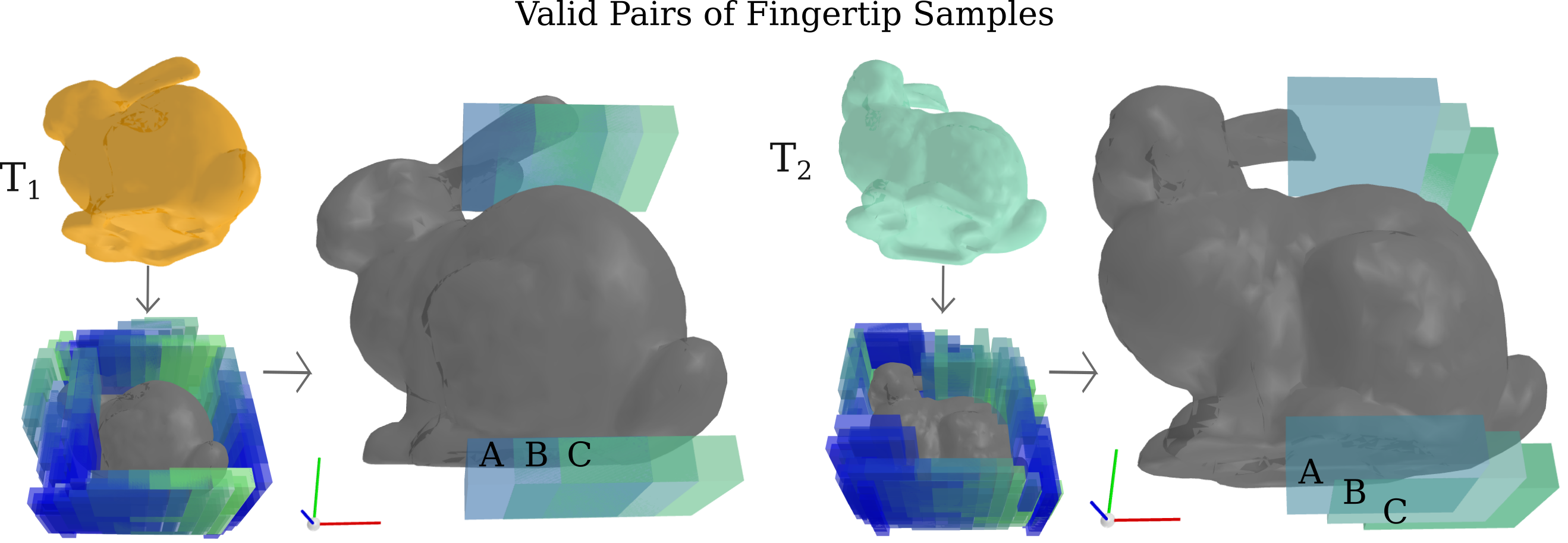}}\\
\subfloat[]{\includegraphics[height=4.2cm]{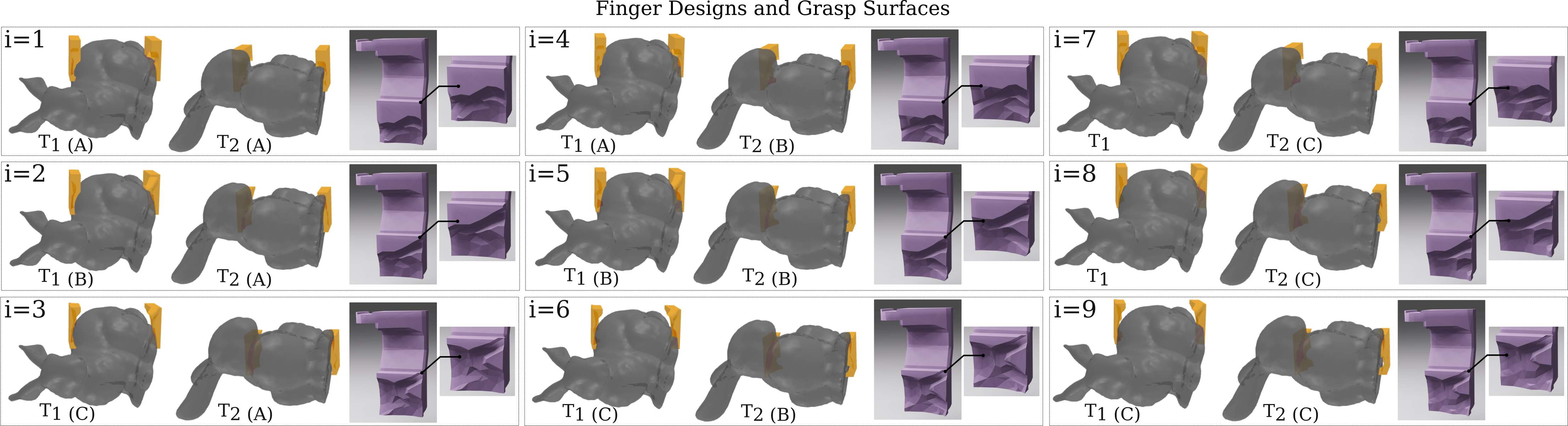}}\\
\end{tabular}
\caption{\textbf{Execution of FCSO} a)~Stable pose generator (Section~\ref{ssec:stablepose}) returned four placements of the bunny and the second and fourth poses were randomly selected; (b)~Grasp sampler (Section~\ref{ssec:graspsampler}) returns three valid pairs of grasp surfaces (A, B, and C) for each pose. $T_1$ depicts the bunny looking towards the left while $T_2$ shows the bunny looking upwards; (c)~Fingerpad customization (Section~\ref{sec:fingerpad_customization}) at these grasp surfaces returned nine possible customized fingerpads that are shown in orange. The gripper fingers (Section~\ref{ssec:fingerdesign}) obtained are shown in purple with the corresponding pose and grasp surface combinations.}%
\label{fig:bunny_evaluation}%
\end{figure*}
We use the Stanford bunny object~\cite{bunny} to evaluate our geometric grasp quality measure with the following parameters:

\begin{itemize}
    \item Robotiq Hand-E gripper (linear opening of 50mm) and its default flat fingers.
    \item Sampling was conducted with a stride $L/5$ and $S$ has dimensions $L=20, W=20, T=5, D=4$.
    \item Two stable placements $(N_p=2)$ with $T_1$ and $T_2$ as the second and fourth object pose in Fig.~\ref{fig:bunny_evaluation}a respectively.
    
\end{itemize}

FCSO returned $N_{s,1}=3$ for $T_1$ and $N_{s,2}=3$ for $T_2$ (Fig.~\ref{fig:bunny_evaluation}b). The number of possible customized grippers would be $C=N_{s,1}*N_{s,2}=3*3=9$ (Fig.~\ref{fig:bunny_evaluation}c) which were evaluated using our geometric grasp quality measure. Each gripper would need to achieve geometric constraints at four surfaces (two surfaces per grasp position as shown in Fig.~\ref{fig:geometryanalysis}a). Table~\ref{table:gqeva} shows the corresponding RLES value of $i^{th}$ gripper fingerpad. It also depicts the effective area for the $i^{th}$ fingerpad geometry at the $m^{th}$ stable pose, $E_{i,m}$, and the quality for the $i^{th}$ fingerpad: $Q_i=min(E_{i,T1}, E_{i,T2})$. The best gripper obtained was $i=9$ with the highest $Q$.

\begin{table}[!t]
\renewcommand{\arraystretch}{1.}
\caption{RLES, contact areas of fingerpads and\\grasp quality at two object poses.}
\label{table:gqeva}
\centering
\begin{tabular}{|c|c|c||c|c||c|c|c|}
\hline
 & \multicolumn{2}{c||}{RLES}&\multicolumn{2}{c||}{Contact area ($A$)}&\multicolumn{3}{c|}{Grasp quality}\\
\hline
$i$ & $T_1$ & $T_2$ & $T_1$ & $T_2$ & $E_{i,T1}$ & $E_{i,T2}$ & $Q_i$\\
\hline
1 & 0.4217 & 0.3878 & 102 & 17.9 & 242.2 & 46.2 & 46.2\\
\hline
2 & 0.5079 & 0.4668 & 110 & 103 & 216.8 & 221.1 & 216.8\\
\hline
3 & 0.5387 & 0.4553 & 126 & 130 & 234.3 & 285.9 & 234.3\\
\hline
4 & 0.5333 & 0.3680 & 103 & 25.9 & 193.2 & 70.4 & 70.4\\
\hline
5 & 0.476 & 0.6297 & 124 & 105 & 261.4 & 167.1 & 167.1\\
\hline
6 & 0.4543 & 0.5509 & 149 & 120 & 329.0 & 217.8 & 217.8\\
\hline
7 & 0.6040 & 0.4529 & 92.7 & 42.3 & 153.5 & 93.4 & 93.4\\
\hline
8 & 0.6515 & 0.5826 & 117 & 94.2 & 180.2 & 161.7 & 161.7\\
\hline
9 & 0.6233 & 0.4863 & 146 & 114.8 & 235.7 & 236.1 & \textbf{235.7}\\
\hline
\end{tabular}
\end{table}

From our experiment, we make two observations: (i) the quality measure requires considering the variation of contact normals and contact surface area to be effective; (ii) the measure is reasonably sufficient in determining the grasp quality as the result coincides with our intuition. The variation of contact normals alone may be insufficient as in Table~\ref{table:gqeva}, the best finger design would be $i=1$ after taking the max-min of the RLES at every $i$. By visual inspection, $i=9$ (Fig.~\ref{fig:bunny_evaluation}c) would provide the best geometric constraints due to more contouring details throughout the fingerpad, which coincides with the result from our proposed quality measure. This measure was used to obtain grippers that achieved successful grasps in Section~\ref{ssec:fingerevaluation}. 

\subsection{Evaluation of customized fingers} \label{ssec:fingerevaluation}

We evaluate the grippers from FCSO by conducting actual pick-and-place experiments for three objects: (1) Intricate cube (L30xW30xH30), (2) Stanford bunny (L65xW50xH65), and (3) L-shaped surgical object (L116xW60xH36). The cube and the L-shaped object are \textit{real} samples produced in HP Labs for certain tasks for the industry, while the bunny was also used in~\cite{kaiyu}, which could serve as a good comparison. These objects would be more suitable for our aim, rather than datasets with common household items without customization such as YCB. The geometrical complexity of customized objects produced in additive manufacturing for the industry is also evident.





In all experiments, a Universal Robot (UR5e) executed at $15^{\circ}/s$ joint speed and $10^{\circ}/s^2$ joint acceleration was used together with a Robotiq Hand-E parallel gripper that has a linear opening of 50mm, specifies a grip force of $60N$ and closing speed of $20mm/s$.



Individual pick-and-place experiments for three objects were conducted (video link in Fig.~\ref{fig:demograsp}). Interestingly, our customized fingerpads contain the most distinct geometries of the object that aid in immobilizing the object. Securely grasping the bunny would seem difficult due to convex geometries and large dimensions compared to the gripper opening. Intuitively, the base of the bunny with small contours along the edges would be the best location to grasp. Our grasp sampler indeed returned valid samples along these extrusions and these contours were also present in the gripper. This observation was evident in both the cube and the L-shaped object, where the internal geometries of the cube and the zig-zag portion of the L-shaped object are present in their respective customized grippers.

A more difficult pick-and-place experiment for different objects and resting poses was also conducted. Objects used were the bunny and the L-shape object resting at two different positions (Fig.~\ref{fig:demograsp}). The best gripper returned would intuitively be the combination of the individual-best grippers for both objects and the result matched our intuition, allowing tightly constrained grasps across all objects and their resting positions, illustrating that the ability of FCSO to be sufficiently versatile for objects in a certain print job. As caging grasps are essentially performed based on a geometrical constraint~\cite{caging1, caging2}, the grasp outcome is highly dependent on the geometry of the gripper rather than friction changes~\cite{kaiyu}. Thus, friction analysis was omitted. 

Additionally, as a single symmetrical finger was returned, it indicates that FCSO had actually planned for 8 different geometries. As each object was not symmetrical, one grasp position requires planning for two geometries. Each object was laid on two different positions, and with one grasp position for each position, this meant that FCSO has planned for four geometries per object. This illustrates the potential of FCSO, as these 8 geometries could had been 8 symmetrical objects.

\begin{figure*}%
\centering\begin{tabular}{c}
\subfloat[]{\includegraphics[height=3.8cm]{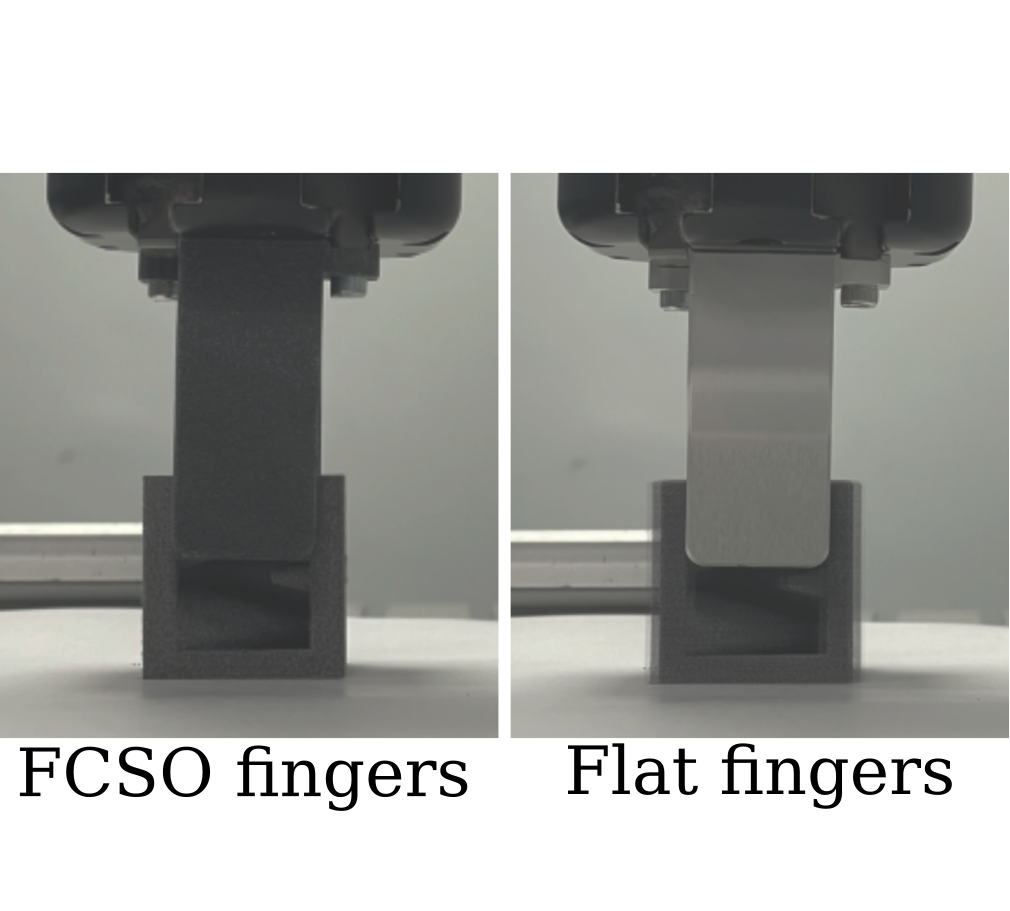}}
\subfloat[]{\includegraphics[height=3.8cm]{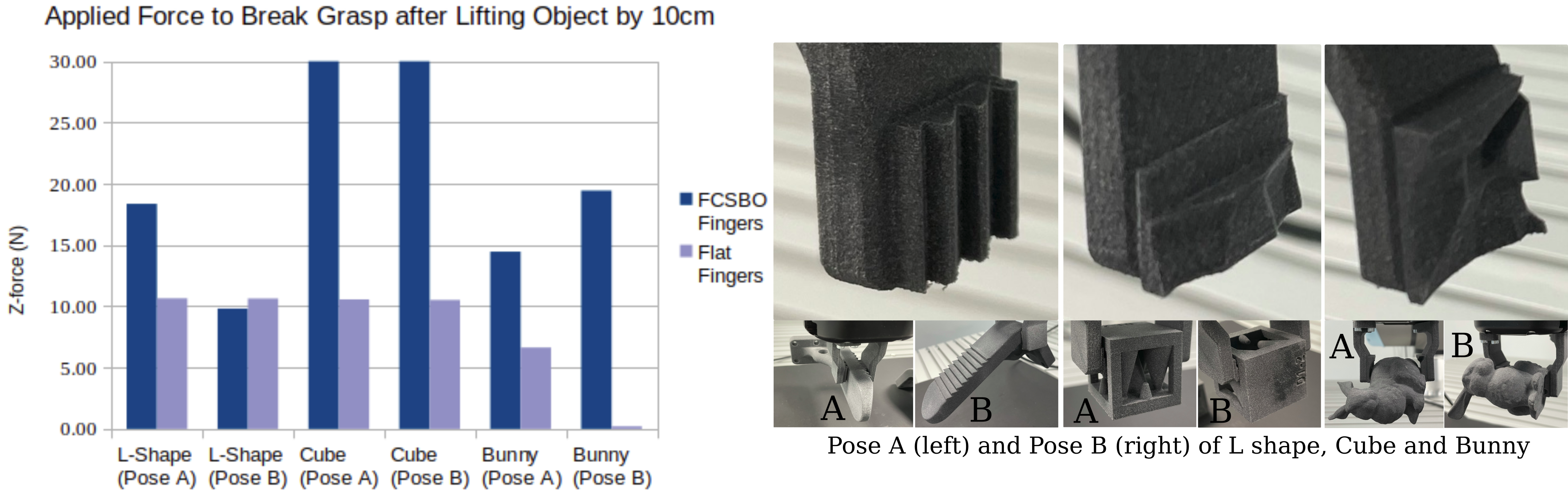}}
\end{tabular}
\caption{\textbf{Precision and stability tests.} Note that all fingers including the flat fingers, were printed using Nylon material (PA11/12) even though a color difference is present. a)~Precision test: The cube was rotated between $-3^{\circ}$ to $3^{\circ}$ before attempting 10 grasps using FCSO fingers and flat fingers. Superimposed images of the cube after grasping showed that position was constant using FCSO fingers while there were positioning errors (shadow) using flat fingers; (b)~Stability test: Printed flat fingers and FCSO fingers were used to grasp and lift objects upwards for 10cm before applying a downward force (maximum $30N$) on the objects. Note that grasps were not broken for both cube poses and the bunny at Pose B slipped out of grasp during the lift for flat fingers.}%
\label{fig:precision_stability}%
\end{figure*}

Objects were manually placed without pose refinement to show that our customized grippers are robust to marginal position errors and uncertainty. Precise positioning can be obtained as the objects slide into contours of the gripper that conform to their geometries during grasps, as evident in Fig.~\ref{fig:precision_stability}a. We also did 10 insertion experiments after grasping for both FCSO fingers and flat fingers, by inserting the 30.4mm cube into a 30.7mm hole without any pose refinement. For the FCSO fingers, we achieved a $100\%$ insertion success rate, while for the flat fingers, only a $10\%$ insertion success rate was observed. This also showed that FCSO fingers are capable of high precision, as the initial pose of the cube was subjected to a positioning error between $-2mm$ and $2mm$ and a rotation error between $-3^{\circ}$ to $3^{\circ}$, yet successful insertion could be achieved for such a tight hole that only has 0.3mm allowance. In addition, we evaluated the holding force to show the stability of the grasps of FCSO fingers against flat fingers in Fig.~\ref{fig:precision_stability}b. Thus, from these experiments, we demonstrated that FCSO could design gripper fingerpads that can achieve precise and stable grasps, which could be implemented in automation for tasks such as sorting or packing.

\section{Conclusion}

The rise in additive manufacturing comes with unique opportunities and challenges. Massive part customization and rapid design changes are made possible with additive manufacturing, however, manufacturing industries that desire the implementation of robotics automation to improve production efficiency could face challenges in the gripper design and grasp planning due to highly complex geometrical shapes resulting from massive customization. Current methods to design robot grippers could be by manual ad-hoc design intuition or automation, which are limiting as the grippers produced from previous methods lack sufficient versatility for practical implementation because they tend to be designed for one object per grasp pose.

Thus, due to the challenge posed by massive customization in additive manufacturing, there is a need for a robust and principled method that can automatically design grippers for 3DP customized and complex objects with sufficient versatility so that automated tasks can be practically executed. Hence, we introduce a fast end-to-end approach that automatically customizes optimal grippers that can grasp different objects at multiple grasping points when given a set of CAD models. To evaluate the grasp quality, we focus on the geometric grasp quality of the contact surfaces based on caging grasps and immobilization. Our geometric grasp quality measure has shown to be reasonably sufficient in differentiating good grippers. We also demonstrated that the designed grippers can grasp multiple objects at different resting poses and are robust to marginal position errors as objects slide into conforming contours of the gripper.

A possible limitation could be the number of objects and scenarios that can be considered. Many objects or positions could lead to over-subtracting of geometries, which may result in relatively flat fingerpads. Although we have not addressed the potential number of objects or geometries that can be used, we showed that it is possible to have 8 objects. In addition, this situation would not occur in mass production where the same set of objects are repeatedly printed. In addition, our idea is to automatically obtain an optimal gripper for a set of 3DP parts in a print job so that grasping and manipulation of these parts could be executed for tasks such as automated sorting or packing. As such, the customized grippers would not be required to exhibit the same degree of versatility as soft grippers, but have the advantage of estimating the pose of the object in-hand as the objects would slide into the contours of these customized grippers. 

\section*{Acknowledgments}
This research was conducted in collaboration with HP Inc. and supported by National Research Foundation (NRF) Singapore and the Singapore Government through the Industry Alignment Grant (I1801E0028).


\bibliographystyle{IEEEtran}
\bibliography{IEEEabrv, ref}

\end{document}